\documentclass[graybox]{svmult}

\usepackage{amsbsy,amssymb,amsmath,amsfonts,latexsym,multicol,multirow,epsfig,color,subfigure,setspace,array,hyperref,url,mathrsfs,bm,algpseudocode,algorithm}

\usepackage{mathptmx}       
\usepackage{helvet}         
\usepackage{courier}        
\usepackage{type1cm}        
\usepackage{makeidx}         
\usepackage{graphicx}        
\usepackage{multicol}        
\usepackage[bottom]{footmisc}

\newcommand{\RR}{\mathbb{R}}
\newcommand{\x}{\bm{x}}
\newcommand{\y}{\bm{y}}
\newcommand{\p}{\bm{p}}
\newcommand{\q}{\bm{q}}
\newcommand{\X}{\mathbf{X}}
\newcommand{\Y}{\mathbf{Y}}
\newcommand{\A}{\mathbf{A}}
\newcommand{\B}{\mathbf{B}}
\newcommand{\C}{\mathbf{C}}
\newcommand{\K}{\mathbf{K}}
\newcommand{\R}{\mathbf{R}}
\newcommand{\bS}{\mathbf{S}}
\newcommand{\bD}{\mathbf{D}}
\newcommand{\bP}{\mathbf{P}}
\newcommand{\independent}{\mathrel{\text{{$\perp\mkern-10mu\perp$}}}}
\newcommand{\bxi}{\bm{\xi}}
\newcommand{\bzeta}{\bm{\zeta}}
\def\T{{\intercal}}

\makeindex             

\begin{document}

\title*{A Review on Modern Computational Optimal Transport Methods with Applications in Biomedical Research}
\author{Jingyi Zhang, Wenxuan Zhong, and Ping Ma}
\institute{Jingyi Zhang \at Center for Statistical Science, Tsinghua University, Beijing, China, \email{joyeecat@gmail.com}
\and Wenxuan Zhong \at Department of Statistics, University of Georgia, Athens, Georgia, USA, \email{wenxuan@uga.edu}
\and Ping Ma \at Department of Statistics, University of Georgia, Athens, Georgia, USA, \email{pingma@uga.edu}}

%
%
\maketitle
\abstract{Optimal transport has been one of the most exciting subjects in mathematics, starting from the 18th century.
As a powerful tool to transport between two probability measures, optimal transport methods have been reinvigorated nowadays in a remarkable proliferation of modern data science applications.
To meet the big data challenges, various computational tools have been developed in the recent decade to accelerate the computation for optimal transport methods.
In this review, we present some cutting-edge computational optimal transport methods with a focus on the regularization-based methods and the projection-based methods. 
We discuss their real-world applications in biomedical research.
}




\section{Introduction}

\begin{figure}[h]
    \begin{center}
        \begin{tabular}{cc}
            \includegraphics[width=0.95\textwidth]{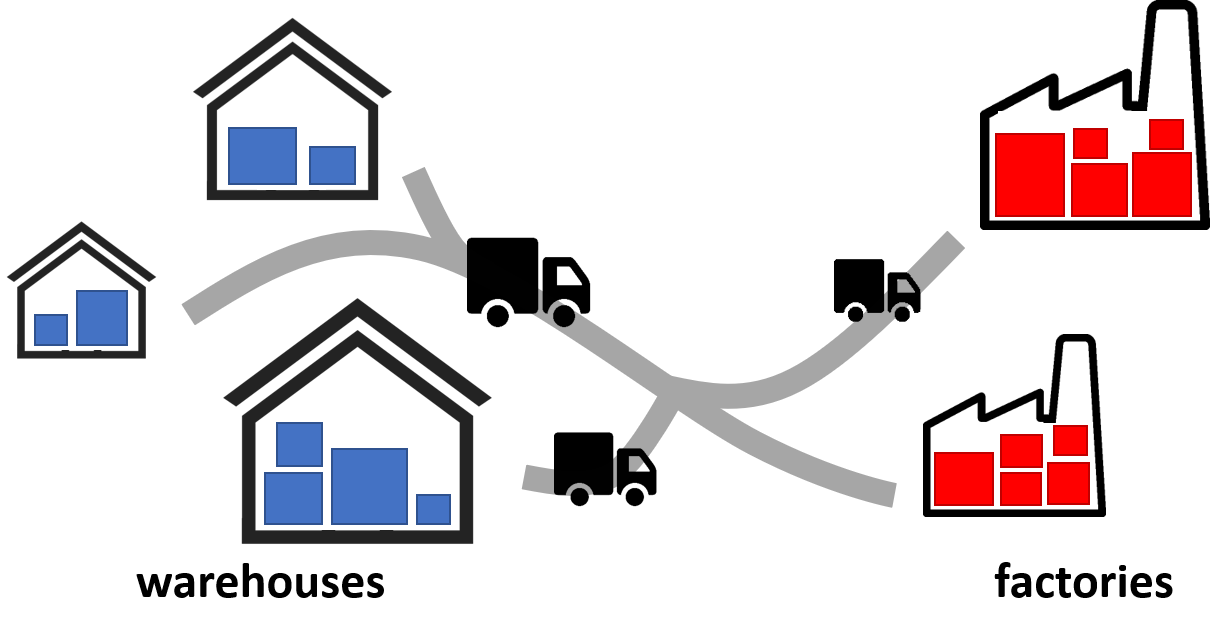}
        \end{tabular}
        \caption{Illustration for the resource allocation problem. The resources in warehouses are marked in blue, and the demand for each factory is marked in red. 
        }\label{fig:1}
    \end{center}
    \vspace*{-0.2in}
\end{figure}

There is a long and rich history of optimal transport (OT) problems initiated by Gaspard Monge (1746–1818), a French mathematician, in the 18th century. 
During recent decades, OT problems have found fruitful applications in our daily lives \cite{villani2008optimal}.
Consider the resource allocation problem, as illustrated in Fig.~\ref{fig:1}.
Suppose that an operator runs $n$ warehouses and $m$ factories. 
Each warehouse contains a certain amount of valuable raw material, i.e., the resources, that is needed by the factories to run properly. 
Furthermore, each factory has a certain demand for raw material.
Suppose the total amount of the resources in the warehouse equals the total demand for the raw material in the factories.
The operator aims to move all the resources from warehouses to factories, such that all the demands for the factories could be successfully met, and the total transport cost is as small as possible.

The resource allocation problem is a typical OT problem in practice. 
To put these problems in mathematical language, one can regard the resources as a whole and the demands as a whole as two probability distributions.
For example, the resources from warehouses in Fig.~1 can be regarded as a non-uniform discrete distribution supported on three discrete points, and each of the points represents the geographical location of a particular warehouse.
OT methods aim to find a transport map (or plan), between these two probability distributions with the minimum transport cost.  
Formal definitions for the transport map, the transport plan, and the transport cost will be given in Section~2.

Nowadays, many modern statistical and machine learning problems can be recast as finding the optimal transport map (or plan) between two probability distributions.
For example, domain adaptation \cite{muzellec2019subspace,courty2017optimal,flamary2019concentration}, aims to learn a well-trained model from a source data distribution and transfer this model to adopt a target data distribution.
Another example is deep generative models \cite{goodfellow2014generative,meng2019large,arjovsky2017wasserstein,chen2018optimal} target at mapping a fixed distribution, e.g., the standard Gaussian or uniform distribution, to the underlying population distribution of the genuine sample. 
During recent decades, OT methods have been reinvigorated in a remarkable proliferation of modern data science applications, including machine learning \cite{alvarez2018structured, courty2017optimal,peyre2019computational, arjovsky2017wasserstein, canas2012learning, flamary2018wasserstein,meng2019large}, statistics \cite{del2019central, cazelles2018geodesic,panaretos2019statistical}, and computer vision \cite{ferradans2014regularized,rabin2014adaptive, su2015optimal,peyre2019computational}.

Although OT finds a large number of applications in practice, the computation of OT meets challenges in the big data era. 
Traditional methods estimate the optimal transport map (OTM) by solving differential equations \cite{brenier1997homogenized,benamou2002monge} or by solving a problem of linear programming \cite{rubner1997earth,pele2009fast}.
Consider two $p$-dimensional samples with $n$ observations within each sample.
The calculation of the OTM between these two samples using these traditional methods requiring $O(n^3\log(n))$ computational time 
\cite{peyre2019computational,seguy2017large}.
Such a sizable computational cost hinders the broad applicability of optimal transport methods.

To alleviate the computational burden for OT, there has been a large number of work dedicated to developing efficient computational tools in the recent decade. 
One class of methods, starting from \cite{cuturi2013sinkhorn}, considers solving a regularized OT problem instead of the original one.
By utilizing the Sinkhorn algorithm (detailed in Section~3), the computational cost for solving such a regularized problem can be reduced to $O(n^2\log(n))$, 
which is a significant reduction from $O(n^3\log(n))$.
Based on this idea, various computational tools are developed to solve the regularized OT problem as quickly as possible \cite{altschuler2017near,peyre2019computational}.
By combining the Sinkhorn algorithm and the idea of low-rank matrix approximation, recently, \cite{altschuler2019massively} proposed an efficient algorithm with a computational cost that is approximately proportional to $n$.
Although not covered in this paper, regularization-based optimal transport methods even appear to have better theoretical properties than the unregularized counterparts; see \cite{genevay2017learning,montavon2016wasserstein,rigollet2018entropic} for details.

Another class of methods aims to estimate the OTM efficiently using random or deterministic projections.
These so-called projection-based methods tackle the problem of estimating a $p$-dimensional OTM 
by breaking down the problem into a series of subproblems, each of which finds a one-dimensional OTM using projected samples \cite{pitie2005n,pitie2007automated,bonneel2015sliced,rabin2011wasserstein}.
The subproblems can be easily solved since the one-dimensional OTM is equivalent to sorting, under some mild conditions.
The projection-based methods reduce the computational cost for calculating OTMs from $O(n^3\log(n))$ to $O(Kn\log(n))$, where $K$ is the number of iterations until convergence.

With the help of these computational tools, OT methods have been widely applied to various biomedical research.
Take single-cell RNA sequencing data as an example, OT methods can be used to study developmental time courses to infer ancestor-descendant fates for cells and help researchers to better understand the molecular programs that guide differentiation during development.
For another example, OT methods can be used as data augmentation tools for increasing the number of observations; and thus to improve the accuracy and the stability of various downstream analyses.

The rest of the paper is organized as follows. 
We start in Section~2 by introducing the essential background of the OT problem.
In Section~3, we present the details of regularization-based OT methods and their extensions.
Section~4 is devoted to projection-based OT methods, including both random projection methods and deterministic projection methods.
In Section~5, we show several applications of OT methods on real-world problems in biomedical research.

\section{Background of the Optimal Transport Problem}

In the aforementioned resource allocation problem, the goal is to transport the resources in the warehouse to the factories with the least cost, say the total fuel consumption of trucks. 
Here, the resources in the warehouse and the demand in the factories can be regarded as discrete distributions.
We now introduce the following example that extend the discrete setting to the continuous setting.  
Suppose there is a worker who has to move a large pile of sand using a shovel in his hand. 
The goal of the worker is to erect with all that sand a target pile with a prescribed shape, say a sandcastle. 
Naturally, the worker wishes to minimize the total ``effort", which intuitively, in the sense of physical, can be regarded as the ``work", the product of force and displacement.
A French mathematician Gaspard Monge (1746–1818) once considered such a problem and formulated it into a general Mathematical problem, i.e., the optimal transport problem \cite{villani2008optimal,peyre2019computational}:
among all the possible transport maps $\phi$ between two probability measures $\mu$ and $\nu$, how to find the one with the minimum transport cost? 
Mathematically, the optimal transport problem can be formulated as follows.
Let $\mathscr{P}(\RR^p)$ be the set of Borel probability measures in $\RR^p$, and let
\begin{eqnarray*}
\mathscr{P}_2(\RR^p)=\left\{\mu\in\mathscr{P}(\RR^p)\Big|\int||x||^2\mbox{d}\mu(x)<\infty\right\}.
\end{eqnarray*}
For $\mu,\nu\in\mathscr{P}_2(\RR^p)$, let $\Phi$ be the set of all the so-called measure-preserving maps $\phi:\mathbb{R}^p\rightarrow\mathbb{R}^p$, such that $\phi_{\#}(\mu) = \nu$ and $\phi^{-1}_{\#}(\nu) = \mu.$
Here, $\#$ represents the push-forward operator, such that for any measurable $\Omega\subset \RR^p$,  $\phi_{\#}(\mu)(\Omega)=\mu(\phi^{-1}(\Omega))$.
Among all the maps in $\Phi$, the optimal transport map defined under a cost function $c(\cdot,\cdot)$ is
\begin{eqnarray}\label{Monge_map}
\phi^\dagger :=\underset{\phi \in \Phi}{\mbox{arg inf}} \int_{\RR^p} c(x, \phi(x)) \mbox{d}\mu(x).
\end{eqnarray}
One popular choice for the cost function is $c(x,y)=\|x-y\|^2$, with which Equation~(\ref{Monge_map}) becomes
\begin{eqnarray}\label{Monge_map2}
\phi^\dagger :=\underset{\phi \in \Phi}{\mbox{arg inf}} \int_{\RR^p} \|x-\phi(x)\|^2 \mbox{d}\mu(x).
\end{eqnarray}

\begin{figure}[b]
\sidecaption
\includegraphics[scale=.45]{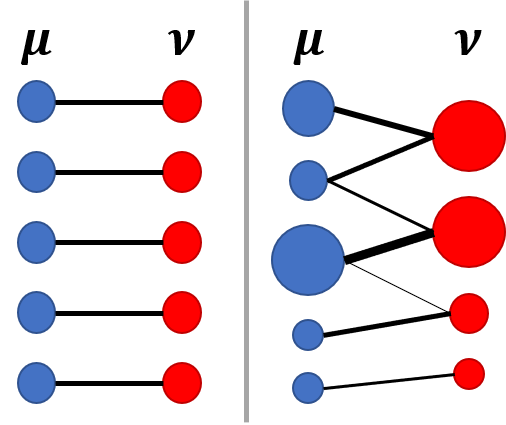}
\caption{Comparison between optimal transport map (OTM) and optimal transport plan (OTP). Left: an illustration of OTM, which is a one-to-one map. Right: an illustration of OPT, which may not necessarily to be a map.}
\label{fig:2}   
\end{figure}

Equation~(\ref{Monge_map2}) is called the Monge formulation, and its solution $\phi^\dagger$ is called the optimal transport map (OTM), or the Monge map. 
The well-known Brenier’s Theorem \cite{brenier1991polar} stated that, when the cost function $c(x,y)=\|x-y\|^2$, if at least one of $\mu$ and $\nu$ has a density with respect to the Lebesgue measure, then the OTM $\phi^\dagger$ in Equation~(\ref{Monge_map2}) exists and is unique.
In other words, the OTM $\phi^\dagger$ may not exists, i.e., the solution of Equation~(\ref{Monge_map2}) may not be a map, when the conditions of Brenier's Theorem is not met.
To overcome such a limitation, Kantorovich \cite{kantorovich1942translation} considered the following set of ``couplings",
\begin{eqnarray}\label{eqn:coupling}
&\mathcal{M}(\mu,\nu)=\{\pi\in\mathscr{P}(\RR^p\times\RR^p) \mbox{ } s.t.\mbox{ } \forall\mbox{ } \text{Borel set} \mbox{ } A, B\subset\RR^p ,\nonumber\\ &\pi(A\times\RR^p)=\mu(A),\mbox{ } \pi(\RR^p\times B)=\nu(B) \}.
\end{eqnarray}
Intuitively, a coupling $\pi\in\mathcal{M}(\mu,\nu)$ is a joint distribution of $\mu$ and $\nu$, such that two particular marginal distributions of $\pi$ are equal to $\mu$ and $\nu$, respectively.
Instead of finding the OTM, Kantorovich formulated the optimal transport problem as finding the optimal coupling,
\begin{eqnarray}\label{K_form}
\pi^* := \underset{\pi \in \mathcal{M}(\mu, \nu)}{\mbox{arg inf}} \int \|x- y \|^2 \mbox{d}\pi(x,y).
\end{eqnarray}
Equation~(\ref{K_form}) is called the Kantorovich formulation (with $L_2$ cost) and its solution $\pi^*$ is called the optimal transport plan (OTP).
The key difference between the Monge formulation and the Kantorovich formulation is that the latter does not require the solution to be a one-to-one map, as illustrated in Fig.~\ref{fig:2}.

The Kantorovich formulation is more realistic in practice, compared with the Monge formulation.
Take the resource allocation problem as an example, as described in Section~1.
It is unreasonable to assume that there always exists a one-to-one map between warehouses and factories, which can meet all the demands for the factories.
The optimal solution of such resource allocation problems thus is usually an OTP instead of an OTM.
Note that, although the Kantorovich formulation is more flexible than the Monge formulation, it can be shown that when the OTM exists, the OTP is equivalent to the OTM. 

Close related to the optimal transport problem is the so-called Wasserstein distance.
Intuitively, if we think the optimal transport problem (either in the Monge formulation or the Kantorovich formulation) as an optimization problem, then the Wasserstein distance is simply the optimal objective value of such an optimization problem, with certain power transform. 
Suppose the OTM $\phi^\dagger$ exists, the Wasserstein distance of order $k$ is defined as
\begin{eqnarray}\label{W_dist}
W_k(\mu, \nu):=\left(\int_{\RR^p} \|X- \phi^\dagger(X) \|^k \mbox{d}\mu \right)^{1/k}.
\end{eqnarray}
Let $\{\x_i\}_{i=1}^n$ and $\{\y_i\}_{i=1}^n$ be two samples generated from $\mu$ and $\nu$, respectively. 
One thus can estimate $\phi^\dagger$ using these two samples, and we let $\widehat{\phi}^\dagger$ to denote the corresponding estimator. 
The Wasserstein distance $W_k(\mu, \nu)$ thus can be estimated by
\begin{eqnarray*}
\widehat{W}_k(\mu, \nu):= \left(\frac{1}{n} \sum\limits_{i=1}^n \|\x_i- \widehat{\phi}^\dagger(\x_i) \|^k  \right)^{1/k}.
\end{eqnarray*}
The Wasserstein distance respecting to the Kantorovich formulation can be defined analogously.
We refer to \cite{weed2019sharp,del2019central2,panaretos2019statistical} and the reference therein for theoretical properties of Wasserstein distances.
Without further notification, we focus on the $L_2$ norm throughout this paper, i.e., $k=2$ in Equation~(\ref{W_dist}), and we abbreviate $W_2(\mu,\nu)$ by $W(\mu,\nu)$.

\section{Regularization-based Optimal Transport Methods}

In this section, we introduce a family of numerical schemes to approximate solutions to the Kantorovich formulation~(\ref{K_form}). 
Such numerical schemes add a regularization penalty to the original optimal transport problem, and one can then solve the regularized problem instead.
Such a regularization-based approach has long been studied in nonparametric regression literature to balances the trade-off between the goodness-of-fit and the model and the roughness of a nonlinear function \cite{gu2013smoothing,ma2015efficient,zhang2018statistical,meng2020more}.

Cuturi first introduced the regularization approach in OT problems \cite{cuturi2013sinkhorn} and showed that the regularized problem could be solved using a simple alternate minimization scheme, requiring $O(n^2\log(n)p)$ computational time. 
Moreover, it can be shown that the solution to the regularized OT problem can well-approximate the solution to its unregularized counterpart.
We call such numerical schemes the regularization-based optimal transport methods.
We now present the details and some extensions of these methods as follows.

\subsection{Computational Cost for OT Problems}
We first introduce how to calculate the empirical Wasserstein distance by solving a linear system.
Let $\p$ and $\q$ be two probability distributions supported on a discrete set $\{\x_i\}_{i=1}^n$, where $\x_i\in\Omega$ for $i=1,\ldots,n$, and $\Omega\subset\RR^p$ is bounded.
We identify $\p$ and $\q$ as the vectors located on the simplex
$$\Delta_n := \left\{\bm{v} \in \RR^n: \mbox{ }\sum_{i=1}^n \bm{v}_i = 1, \mbox{ and }\bm{v}_i\geq0, \mbox{ } i=1,\ldots,n. \right\},$$
whose entries denote the weight of each distribution assigned to the points of $\{\x_i\}_{i=1}^n$.
Let $\C\in\RR^{n\times n}$ be the pair-wise distance matrix, where $\C_{ij} = \|\x_i-\x_j\|^2$, and $\mathbf{1}_n$ be the all-ones vector with $n$ elements.
Recall the definition of coupling in Equation~(\ref{eqn:coupling}), and analogously, we denote by $\mathcal{M}(\p, \q)$ the set of coupling matrices between $\p$ and $\q$, i.e.,
\begin{eqnarray*}
\mathcal{M}(\p, \q)=\left\{ \bP\in\RR^{n\times n}:\mbox{ } \bP\mathbf{1}_n=\p,\mbox{ } \bP^\T\mathbf{1}_n=\q \right\}.
\end{eqnarray*}
For brevity, this paper focuses on
square matrices $\C$ and $\bP$, since extensions to rectangular cases are straightforward.

Let $\langle \cdot,\cdot \rangle$ denote the summation of the element-wise multiplication, such that, for any two matrix $\mathbf{A},\mathbf{B}\in\RR^{n\times n}$, $\langle \mathbf{A},\mathbf{B} \rangle=\sum_{i=1}^n\sum_{j=1}^n \mathbf{A}_{ij}\mathbf{B}_{ij}$.
According to the Kantorovich formulation in Equation~(\ref{K_form}), 
the Wasserstein distance between $\p$ and $\q$, i.e., $W(\p, \q)$ thus can be calculated through solving the following optimization problem
\begin{eqnarray}\label{OT_dist}
\underset{\bP \in \mathcal{M}(\p,\q)}{\min} \left \langle \bP,\C \right\rangle,
\end{eqnarray}
which is a linear program with $O(n)$ linear constraints.
The coupling matrix $\bP$ is called the optimal coupling matrix, when the optimization problem~(\ref{OT_dist}) achieves the minimum value, i.e., the optimal coupling matrix is the minimizer of the optimization problem~(\ref{OT_dist}).
Note that when the OTM exists, the optimal coupling matrix $\bP$ is a sparse matrix, such that there is exactly one non-zero element in each row and each column of $\bP$, respectively.

Practical algorithms for solving the problem~(\ref{OT_dist}) through linear programming requiring a computational time of the order $O(n^3 \log(n))$ for fixed $p$ \cite{peyre2019computational}.
Such a sizable computational cost hinders the broad applicability of OT methods in practice for the datasets with large sample size.

\subsection{Sinkhorn Distance}
To alleviate the computation burden for OT problems, \cite{cuturi2013sinkhorn} considered a variant of the minimization problem in Equation~(\ref{OT_dist}), which can be solved within $O(n^2\log(n)p)$ computational time using the Sinkhorn scaling algorithm, originally proposed in \cite{sinkhorn1967diagonal}.
The solution of such a variant is called the Sinkhorn ``distance" \footnote{We use quotations here since it is not technically a distance; see Section 3.2 of \cite{cuturi2013sinkhorn} for details. The quotes are
dropped henceforth.}, defined as 
\begin{eqnarray}\label{sink_dist}
W_\eta(\p, \q) = \underset{\bP \in \mathcal{M}(\p,\q)}{\min} \left \langle \bP,\C \right\rangle-\eta^{-1}H(\bP),
\end{eqnarray}
where $\eta>0$ is the regularization parameter, and $H(\bP)=\sum_{i=1}^n\sum_{j=1}^n\bP_{ij}\log(1/\bP_{ij})$ is the Shannon entropy of $\bP$.
We adopt the standard convention that $0\log(1/0) = 0$ in the Shannon entropy.
We present a fundamental definition as follows \cite{sinkhorn1967diagonal}.
\begin{definition}
Given $\p$, $\q \in \Delta_n$ and $\K\in \RR^{n\times n}$ with positive entries, the Sinkhorn projection $\Pi_{\mathcal{M}(\p,\q)} (\K)$ of $\K$ onto $\mathcal{M}(\p, \q)$ is the unique matrix in $\mathcal{M}(\p, \q)$ of the form $\bD_1\K\bD_2$ for positive diagonal matrices $\bD_1,\bD_2\in\RR^{n\times n}$.
\end{definition}

\begin{figure}[h]
    \begin{center}
        \begin{tabular}{cc}
            \includegraphics[width=0.95\textwidth]{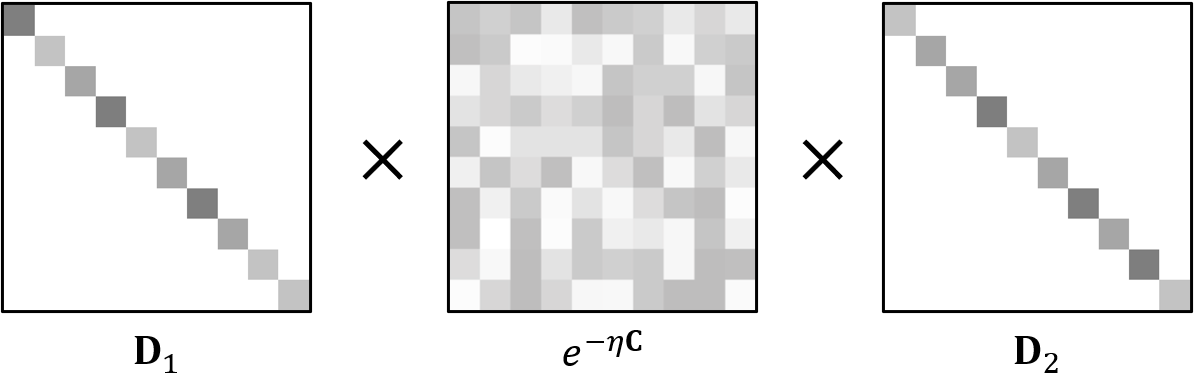}
        \end{tabular}
        \caption{The minimizer of the regularized optimal transport problem~\eqref{sink_dist} takes the form $\bD_1(e^{-\eta\C})\bD_2$, for some unknown diagonal metrics $\bD_1$ and $\bD_2$. }\label{fig:sinksolve}
    \end{center}
    \vspace*{-0.2in}
\end{figure}

Let $\bP^\eta$ be the minimizer, i.e., the optimal coupling matrix, of the optimization problem~(\ref{sink_dist}).
Throughout the paper, all matrix exponentials and logarithms will be taken entrywise, i.e., $(e^\mathbf{A})_{ij} :=e^{\mathbf{A}_{ij}}$ and $(\log \mathbf{A})_{ij} := \log \mathbf{A}_{ij}$ for any matrix $\mathbf{A} \in\RR^{n\times n}$.
\cite{cuturi2013sinkhorn} built a simple but key connection between the Sinkhorn distance and the Sinkhorn projection,
\begin{align}\label{sink_dist2}
\bP^\eta &= \underset{\bP \in \mathcal{M}(\p,\q)}{\mbox{argmin}}\left \langle \bP, \C \right\rangle-\eta^{-1}H(\bP) \nonumber\\
&= \underset{\bP \in \mathcal{M}(\p,\q)}{\mbox{argmin}}\left \langle \eta\C, \bP \right\rangle-\eta^{-1}H(\bP) \nonumber\\
&=\underset{\bP \in \mathcal{M}(\p,\q)}{\mbox{argmin}}\left \langle -\log \left(e^{-\eta\C}\right),\bP \right\rangle-\eta^{-1}H(\bP) \nonumber\\
&= \Pi_{\mathcal{M}(\p,\q)}\left(e^{-\eta\C}\right).
\end{align}
Equation~(\ref{sink_dist2}) suggests the minimizer of the optimization problem~(\ref{sink_dist}) takes the form $\bD_1(e^{-\eta\C})\bD_2$, for some positive diagonal matrices $\bD_1,\bD_2\in\RR^{n\times n}$, as illustrated in Fig.~\ref{fig:sinksolve}.
Moreover, it can be shown that the minimizer in Equation~(\ref{sink_dist2}) exists and is unique due to the strict convexity of $-H(\bP)$ and the compactness of $\mathcal{M}(\p, \q)$.

Based on Equation~(\ref{sink_dist2}), \cite{cuturi2013sinkhorn} proposed a simple iterative algorithm, which is also known as the Sinkhorn-Knopp algorithm, to approximate $\bP^\eta$.
Let $x_i, y_i, p_i, q_i$ be the $i$-th element of the vector $\x,\y,\p$, and $\q$, respectively, for $i=1,\ldots,n$. 
For simplicity, we now use $\A$ to denote the matrix $e^{-\eta\C}$.
Intuitively, the Sinkhorn-Knopp algorithm works as an alternating projection procedure that renormalizes the rows and columns of $\A$ in turn, so that they match the desired row and column marginals $\p$ and $\q$. 
In specific, at each step, it prescribes to either modify all the rows of $\A$ by multiplying the $i$-th row by $(p_i/\sum_{j=1}^n\A_{ij})$, for $i=1,\ldots,n$, or to do the analogous operation on the columns. 
Here, $\sum_{j=1}^n \A_{ij}$ is simply the $i$-th row sum of $\A$.
Analogously, we also use $\sum_{i=1}^n \A_{ij}$ to denote the $j$-th column sum of $\A$.
The standard convention that $0/0 = 1$ is adopted in the algorithm if it occurs.
The algorithm terminates when the matrix $\A$, after $k$-th iteration, is sufficiently close to the polytope $\mathcal{M}(\p,\q)$.
The pseudocode for the Sinkhorn-Knopp algorithm is shown in Algorithm~\ref{alg:ALG3}.

\begin{algorithm}
    \caption{{\sc Sinkhorn($\A, \mathcal{M}(\p,\q),\epsilon$)}}
    \label{alg:ALG3}
    \begin{algorithmic}
        \State \textbf{Initialize:} $k\leftarrow 0$;\enspace $\A^{[0]}\leftarrow \A/\|\A\|_1$;\enspace
        $\x^{[0]}\leftarrow\mathbf{0}$;\enspace $\y^{[0]}\leftarrow\mathbf{0}$\enspace
        \Repeat
        \State $k\leftarrow k+1$
            \State \textbf{if} $k$ is odd \textbf{then}
            \State \qquad $x_i\leftarrow\log(p_i/\sum_{j=1}^n\A^{[k-1]}_{ij})$, \enspace for $i=1,\ldots,n$
            \State \qquad $\x^{[k]}\leftarrow\x^{[k-1]}+\x$;\enspace $\y^{[k]}\leftarrow\y^{[k-1]}$
            \State \textbf{else}   
            \State \qquad $y_j\leftarrow\log(q_i/\sum_{i=1}^n\A^{[k-1]}_{ij})$, \enspace for $j=1,\ldots,n$
            \State \qquad $\y^{[k]}\leftarrow\y^{[k-1]}+\y$;\enspace $\x^{[k]}\leftarrow\x^{[k-1]}$
            \State $\bD_1\leftarrow \mbox{diag}(\exp(\x^{[k]}))$;\enspace
            $\bD_2\leftarrow \mbox{diag}(\exp(\y^{[k]}))$
            \State$\A^{[k]}=\bD_1\A\bD_2$
        \Until $\mbox{dist}(\A^{[k]}, \mathcal{M}(\p,\q))\leq\epsilon$
        \State \textbf{Output:} $\bP^\eta = \A^{[k]}$
    \end{algorithmic}
\end{algorithm}


One question remaining for Algorithm~\ref{alg:ALG3} is how to determine the size of $\eta$, which balances the trade-off between the computation time and the estimation accuracy.
In specific, a small $\eta$ is associated with a more accurate estimation of the Wasserstein distance as well as longer computation time \cite{genevay2019sample}.

Algorithm~\ref{alg:ALG3} requires a computational cost of the order $O(n^2\log(n)pK)$, where $K$ is the number of iterations.
It is known that $K=O(\epsilon^{-2})$ in order to let Algorithm~\ref{alg:ALG3} to achieve the desired accuracy.
Recently, \cite{altschuler2017near} proposed a new greedy coordinate descent variant of the Sinkhorn algorithm with the same theoretical guarantees and a significantly smaller number of iterations.
With the help of Algorithm~\ref{alg:ALG3}, the regularized optimal transport problem can be solved reliably and efficiently in the cases when $n\approx 10^4$ \cite{cuturi2013sinkhorn,genevay2016stochastic}.

\subsection{Sinkhorn Algorithms with the Nystr$\ddot{\mathit{\textbf{o}}}$m Method}

Although the Sinkhorn-Knopp algorithm has already yielded impressive algorithmic benefits, its computational complexity and memory usage are of 
the order of $n^2$, since such an algorithm involves the calculation of the $n\times n$ matrix $e^{-\eta\C}$.
Such a quadratic computational cost makes the calculation of Sinkhorn distances prohibitively expensive on the datasets with millions of observations.

To alleviate the computation burden, \cite{altschuler2019massively} proposed to replace the computation of the entire matrix $e^{-\eta\C}$ with its low-rank approximation.
Computing such approximations is a problem that has long been studied in machine learning under different names, including Nystr$\ddot{\mbox{o}}$m method \cite{williams2001using,wang2013improving}, sparse greedy approximations \cite{smola2000sparse}, incomplete Cholesky decomposition \cite{fine2001efficient}, and CUR matrix decomposition \cite{mahoney2009cur}. 
These methods draw great attention in the subsampling literature due to its close relationship to the {\it algorithmic leveraging} approach \cite{ma2015leveraging,meng2017effective,zhang2018statistical,ma2020asymptotic}, which has been widely applied in linear regression models \cite{mahoney2011randomized,drineas2012fast,ma2015statistical}, logistic regression \cite{wang2018optimal}, and streaming time series \cite{xie2019online}.
Among the aforementioned low-rank approximation methods, the Nystr$\ddot{\mbox{o}}$m method is arguably the most extensively used one in the literature \cite{wang2015practical,mahoney2016lecture}.
We now briefly introduce Nystr$\ddot{\mbox{o}}$m method, followed by the fast Sinkhorn algorithm proposed in \cite{altschuler2019massively} that utilize Nystr$\ddot{\mbox{o}}$m for low-rank matrix approximation.

\begin{figure}[h]
    \begin{center}
        \begin{tabular}{cc}
            \includegraphics[width=0.95\textwidth]{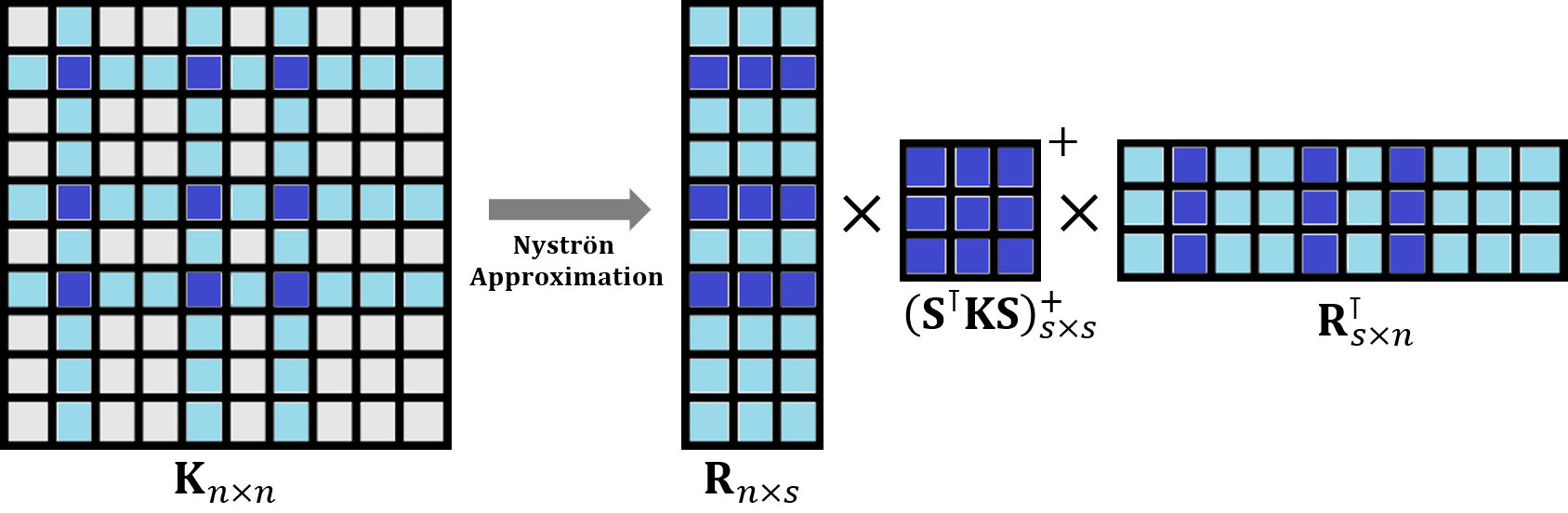}
        \end{tabular}
        \caption{Illustration for the Nystr$\ddot{\mbox{o}}$m method. }\label{fig:nystron}
    \end{center}
    \vspace*{-0.2in}
\end{figure}

Let $\K\in\RR^{n\times n}$ be the matrix that we aim to approximate.
Let $s<n$ be a positive integer, $\bS$ be a $n\times s$ column selection matrix \footnote{A column selection matrix is the one that all the elements of which equals zero except that there exists one element in each column that equals one.}, and $\R=\K\bS\in\RR^{n\times s}$ be the so-called sketch matrix of $\K$.
In other words, $\R$ is a matrix that contains certain columns of $\K$.
Consider the optimization problem
\begin{eqnarray}\label{nystrom}
\widetilde{\X}=\underset{\X\in\RR^{s\times s}}{\mbox{argmin}}\|\bS^\T(\K-\R\X\R^\T)\bS\|^2_F,
\end{eqnarray}
where $\|\cdot\|_F$ denotes the Frobenius norm.
Equation~(\ref{nystrom}) suggests the matrix $\R\widetilde{\X}\R^\T$ can be utilized as a low-rank approximation of $\K$, since such a matrix is the closest one to $\K$ among all the semi-positive definite metrics that have rank at most $s$.
Let $(\cdot)^+$ to denote the Moore-Penrose inverse of a matrix.
It is known that the minimizer of the optimization problem~(\ref{nystrom}) takes the form 
\begin{eqnarray*}
\widetilde{\X} = (\bS^\T\R)^+(\bS^\T\K\bS)(\R^\T\bS)^+ = (\bS^\T\K\bS)^+;
\end{eqnarray*}
see \cite{wang2015practical} for technical details.
Consequently, we have the following low-rank approximation of $\K$,
\begin{eqnarray*}
\K\approx \R(\bS^\T\K\bS)^+\R^\T,
\end{eqnarray*}
and such an approximation is called the Nystr$\ddot{\mbox{o}}$m method, as illustrated in Fig.~\ref{fig:nystron}.
It is known that the Nystr$\ddot{\mbox{o}}$m method is highly efficient, and could reliably be run on problems of size $n\approx 10^6$ \cite{wang2015practical}.

Algorithm~\ref{alg:ALG4} introduces NYS-SINK \cite{altschuler2019massively}, i.e., the Sinkhorn algorithm implemented with the the Nystr$\ddot{\mbox{o}}$m method.
The notations are analogous to the ones in Algorithm~\ref{alg:ALG3}.
\begin{algorithm}
    \caption{{\sc Nys-sink($\A, \mathcal{M}(\p,\q),\epsilon$,$s$)}}
    \label{alg:ALG4}
    \begin{algorithmic}
        \State \textbf{Input:} $\A$, $\p$, $\q$, $s$
        \State \textit{Step 1:} Calculate the Nystr$\ddot{\mbox{o}}$m approximation of $\A$ (with rank $s$), denoted by $\widetilde{\A}.$ 
        \State \textit{Step 2:} $\widetilde{\bP}^\eta$ = {\sc Sinkhorn($\widetilde{\A}, \mathcal{M}(\p,\q),\epsilon$)}
        \State \textbf{Output:} $\widetilde{\bP}^\eta$
    \end{algorithmic}
\end{algorithm}
Algorithm~\ref{alg:ALG4} requires a memory cost of the order $O(ns)$ and a computational cost of the order $O(ns^2p)$. 
When $s\ll n$, these costs are significant reductions compared with $O(n^2)$ and $O(n^2\log(n)p)$ for Algorithm~\ref{alg:ALG3}, respectively.
\cite{altschuler2019massively} reported that Algorithm~\ref{alg:ALG4} could reliably be run on problems of size $n\approx 10^6$ on a single laptop.

There are two fundamental questions when implementing the Nystr$\ddot{\mbox{o}}$m method in practice: (1) how to decide the size of $s$; and (2) given $s$, how to construct the column selection matrix $\bS$.
For the latter question, we refer to \cite{gittens2016revisiting} for an extensive review of how to construct $\bS$ through weighted random subsampling.
There also exists recursive strategy \cite{musco2017recursive} for potentially more effective construction of $\bS$. 
For the former question, various data-driven strategies have been proposed to determine the size of $s$ that is adaptive to the low-dimensional structure of the data.
These strategies are developed under different model setups, including kernel ridge regression \cite{gittens2016revisiting,musco2017recursive,calandriello2020analysis}, kernel K-means \cite{he2018kernel,wang2019scalable}, and so on.
Recently, \cite{an2021efficient} further improved the efficiency through Nesterov's smoothing technique.
Consider the optimal transport problem that of our interest, \cite{altschuler2019massively} assumed the data are lying on a low-dimensional manifold, and the authors developed a data-driven strategy to determine the effective dimension of such a manifold.

\section{Projection-based Optimal Transport Methods}
In the cases when $n\gg p$, one can utilize projection-based optimal transport methods for potential faster calculation as well as smaller memory consumption, compared with regularization-based optimal transport methods.
These projection-based methods build upon a key fact that the empirical one-dimensional OTM under the $L_2$ norm is equivalent to sorting.
Utilizing such a fact, the projection-based OT methods tackle the problem of estimating a $p$-dimensional OTM 
by breaking down the problem into a series of subproblems, each of which finds a one-dimensional OTM using projected samples \cite{pitie2005n,pitie2007automated,bonneel2015sliced,rabin2011wasserstein}.
The projection direction can be selected either at random or at deterministic, based on different criteria. 
Generally speaking, the computational cost for these projection-based methods are approximately proportional to $n$, and the memory cost of which is at the order of $O(np)$, which is a significant reduction from $O(n^2)$ when $p\ll n$.
We will cover some representatives of the projection-based OT methods in this section.

\subsection{Random Projection OT Method}
The random projection method, also called the Radon probability density function (PDF) transformation method, is first proposed in \cite{pitie2005n} for transferring the color between different images.
Intuitively, an image can be represented as a three-dimensional sample in the RGB color space, in which each pixel of the image is an observation.
The goal of color transfer is to find a transport map $\phi$ such that the color of the transformed source image follows the same distribution of the color of the target image.
Although the map $\phi$ does not have to be the OTM in this problem, the random projection method proposed in \cite{pitie2005n} can be regarded as an estimation method for OTM.

\begin{figure}[h]
    \begin{center}
        \begin{tabular}{cc}
            \includegraphics[width=0.95\textwidth]{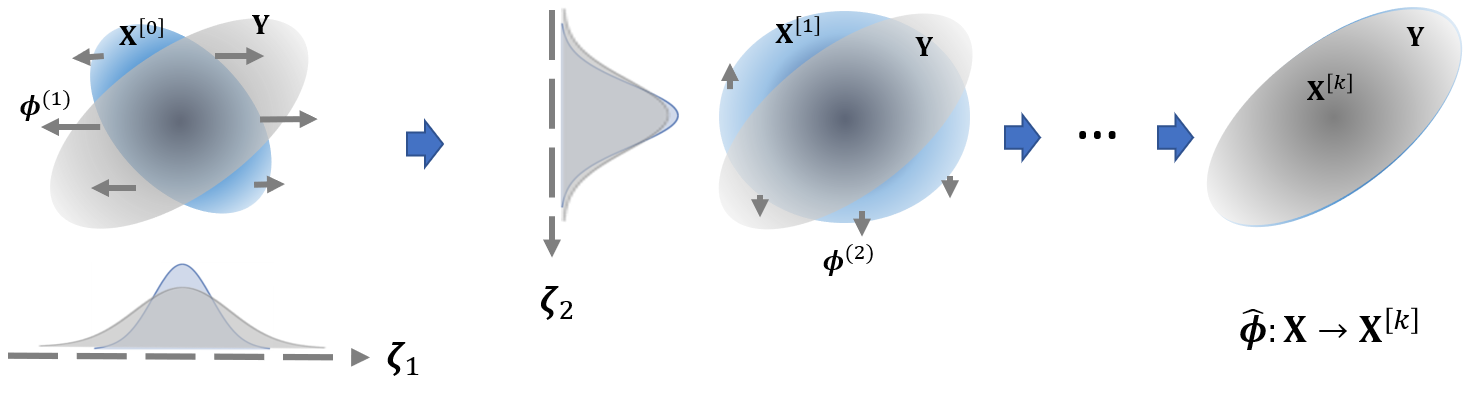}
        \end{tabular}
        \caption{Illustration of Algorithm~\ref{alg:ALG1}. In the $k$-th iteration, a random projection direction $\bzeta_k$ is generated, and the one-dimensional OTM is calculated that match the projected sample $\X^{[k]}\bzeta_k$ to $\Y\bzeta_k$.  }\label{fig:alg1}
    \end{center}
    \vspace*{-0.2in}
\end{figure}

The random projection method is built upon the fact that two PDFs are identical if the marginal distributions, respecting all possible one-dimensional projection directions, of these two PDFs, are identical.
Since it is impossible to consider all possible projection directions in practice, the random projection method thus utilizes the Monte Carlo method and considers a sequence of randomly generated projection directions.
The details of the random projection method are summarized in Algorithm~\ref{alg:ALG1}.
The computational cost for Algorithm~\ref{alg:ALG1} is at the order of $O(n\log(n)pK)$, where $K$ is the number of iterations under converge.
We illustrate Algorithm~\ref{alg:ALG1} in Fig.~\ref{fig:alg1}.

\begin{algorithm}[ht]
\caption{Random projection method for OTM}
        \label{alg:ALG1}
\begin{algorithmic}
\State \textbf{Input:} the source matrix $\X\in \mathbb{R}^{n\times p}$ and the target matrix $\Y\in \mathbb{R}^{n\times p}$
    \State $k\leftarrow 0$,\enspace $\X^{[0]}\leftarrow \X$
    \Repeat
    \State (a) generate a random projection direction $\bzeta_k\in\mathbb{R}^p$
    \State (b) find the one-dimensional OTM $\phi^{(k)}$ that matches $\X^{[k]}\bzeta_k$ to $\Y\bzeta_k$
    \State (c) $\X^{[k+1]}\leftarrow \X^{[k]}+(\phi^{(k)}(\X^{[k]}\bzeta_k)-\X^{[k]}\bzeta_k)\bzeta_k^\T$
    \State (d) $k\leftarrow k+1$
    \Until converge
    \State The final estimator is given by $\widehat{\phi}:\X\rightarrow\X^{[k]}$

\end{algorithmic}
\end{algorithm}

Instead of randomly generating the projection directions using the Monte Carlo method, one can also generate a sequence of projection directions with ``low-discrepancy", i.e., the directions that are distributed as disperse as possible on the unit sphere.
The low-discrepancy sequence has been widely applied in the field of quasi-Monte Carlo and has been extensively employed for numerical integration \cite{owen2003quasi} and subsampling in big data \cite{meng2020more}.
We refer to \cite{lemieux2009book,leobacher2014introduction,dick2013high,glasserman2013monte} for more in-depth discussions on quasi-Monte Carlo methods.
It is reported in \cite{pitie2005n} that using a low-discrepancy sequence of projection directions yields a potentially faster convergence rate.

Close related to the random projection method is the sliced method.
The sliced method modifies the random projection method by considering a large set of random directions from $\mathbb{S}^{d-1}$ in each iteration, where $\mathbb{S}^{d-1}$ is the $d$-dimensional unit sphere.
The ``mean map'' of the one-dimensional OTMs over these random directions is considered as a component of the final estimate of the desired OTM.
Let $L$ be the number of projection directions considered in each iteration.
Consequently, the computational cost of the sliced method is at the order of $O(n\log(n)pKL)$, where $K$ is the number of iterations until convergence.
Although the sliced method is $L$ times slower than the random projection method, in practice, it is usually observed that the former yields a more robust estimation of the latter.
We refer to \cite{bonneel2015sliced,rabin2011wasserstein} for more implementation details of the sliced method.

\subsection{Projection Pursuit OT Method}
Despite the random projection method works reasonably well in practice, for moderate or large $p$, such a method suffers from slow or none convergence due to the nature of randomly selected projection directions. 
To address this issue, \cite{meng2019large} introduced a novel statistical approach to estimate large-scale OTMs \footnote{The code is available at https://github.com/ChengzijunAixiaoli/PPMM.}. 
The proposed method, named projection pursuit Monge map (PPMM), combines the idea of projection pursuit \cite{friedman1981projection} and sufficient dimension reduction \cite{li2018sufficient}.
The projection pursuit technique is similar to boosting that search for the next optimal direction based on the residual of previous ones. 
In each iteration, PPMM aims to find the ``optimal" projection direction, guided by sufficient dimension reduction techniques, instead of using a randomly selected one. 
Utilizing these informative projection directions, it is reported in \cite{meng2019large} that the PPMM method yields a significantly faster convergence rate than the random projection method.
We now introduce some essential background of sufficient dimension reduction techniques, followed by the details of the PPMM method.

Consider a regression problem with a univariate response $T$ and a $p$-dimensional predictor $Z$.
Sufficient dimension reduction techniques aim to reduce the dimension of $Z$ while preserving its regression relation with $T$. 
In other words, such techniques seek a set of linear combinations of $Z$, say $\B^\T Z$ with some projection matrix $\B\in~\RR^{p \times q}$ ($q<p$), such that $T$ depends on $Z$ only through $\B^{\T}Z$, i.e., 
\begin{eqnarray}\label{eqn:sdr}
T \independent Z | \B^{\T}Z.
\end{eqnarray}
Let ${\cal S}(\B)$ to denote the column space of $\B$. 
We call ${\cal S}(\B)$ a sufficient dimension reduction subspace (s.d.r. subspace) if $\B$ satisfy Formulation~(\ref{eqn:sdr}).
Moreover, if the intersection of all possible s.d.r. subspaces is still an s.d.r. subspace, we call it the central subspace and denote it as ${\cal S}_{T|Z}$. 
Note that the central subspace is the s.d.r. subspace with the minimum number of dimensions.
Some popular sufficient dimension reduction techniques include sliced inverse regression (SIR) \cite{li1991sliced}, principal Hessian directions (PHD) \cite{li1992principal}, sliced average variance estimator (SAVE) \cite{cook1991sliced}, directional regression (DR) \cite{li2007directional}, among others. 
Under some regularity conditions, it can be shown that these methods can induce an s.d.r. subspace that equals the central subspace. 

Consider estimating the OTM between a source sample and a target sample. 
One can form a regression problem using these two samples, i.e., add a binary response variable by labeling them as 0 and 1, respectively. 
The PPMM method utilizes sufficient dimension reduction techniques to select the most ``informative'' projection direction.
Here, we call a projection direction $\bxi$ the most informative one, if the projected samples have the most substantial `` discrepancy.'' 
The discrepancy can be measured by the difference of the $k$th order moments or central moments.
For example, the SIR method measures the discrepancy using the difference of means, while the SAVE method measures the discrepancy using the difference of variances.
The authors in \cite{meng2019large} considered the SAVE method and showed that the most informative projection direction was equivalent to the eigenvector corresponding to the largest eigenvalue of the projection matrix $\B$, estimated by SAVE. 
The detailed algorithm for PPMM is summarized in Algorithm \ref{alg:ALG2} as follows.

\begin{algorithm}
        \caption{Projection pursuit Monge map (PPMM)}
        \label{alg:ALG2}
        \begin{algorithmic}
        \State \textbf{Input:} two matrix $\X\in \mathbb{R}^{n\times p}$ and $\Y\in \mathbb{R}^{n\times p}$
    \State $k\leftarrow 0$,\enspace $\X^{[0]}\leftarrow \X$
    \Repeat
    \State (a) calculate the most informative projection direction $\bxi_k\in\mathbb{R}^p$ between $\X^{[k]}$ and $\Y$ using SAVE 
    \State (b) find the one-dimensional OTM $\phi^{(k)}$ that matches $\X^{[k]}\bxi_k$ to $\Y\bxi_k$
    \State (c) $\X^{[k+1]}\leftarrow \X^{[k]}+(\phi^{(k)}(\X^{[k]}\bxi_k)-\X^{[k]}\bxi_k)\bxi_k^\T$
    \State (d) $k\leftarrow k+1$
    \Until converge
    \State The final estimator is given by $\widehat{\phi}:\X\rightarrow\X^{[k]}$
    \end{algorithmic}
\end{algorithm}

\begin{figure}[h]
    \begin{center}
        \begin{tabular}{cc}
            \includegraphics[width=0.95\textwidth]{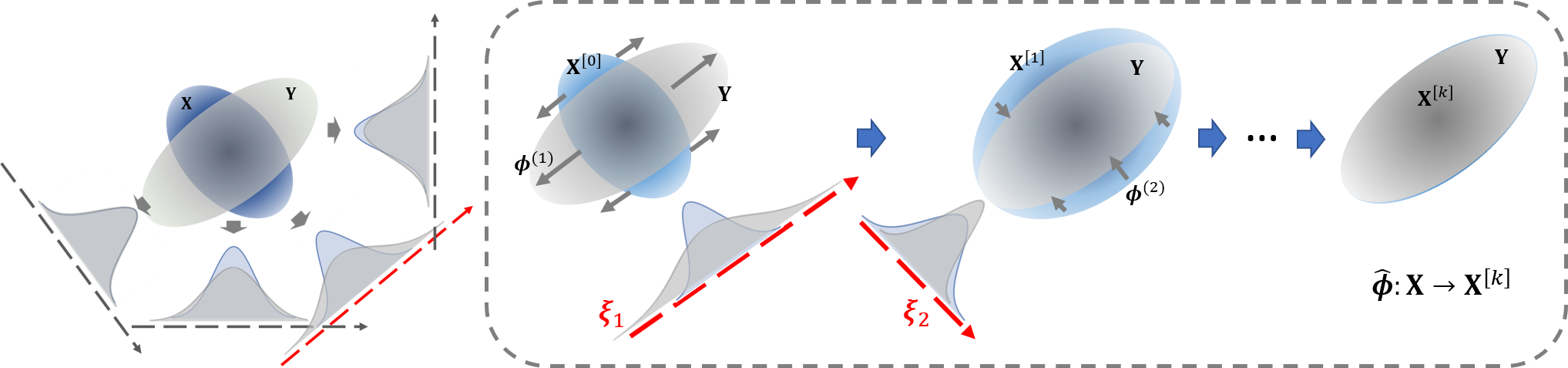}
        \end{tabular}
        \caption{Illustration of Algorithm~\ref{alg:ALG2}. The left panel shows that in the $k$-th iteration, the most informative projection direction $\bxi_k$ is calculated by SAVE. The right panel shows that the one-dimensional OTM is calculated to match the projected sample $\X^{[k]}\bxi_k$ to $\Y\bxi_k$. }\label{fig:alg2}
    \end{center}
    \vspace*{-0.2in}
\end{figure}

The computational cost for Algorithm~\ref{alg:ALG2} mainly resides in steps (a) and (b). 
Within each iteration, steps (a) and (b) require the computational cost of the order $O(np^2)$ and $O(n\log(n)$, respectively. 
Consequently, the overall computational cost for Algorithm~\ref{alg:ALG2} is at the order of $O(Knp^2+Kn\log(n))$, where $K$ is the number of iterations.
Although not theoretical guaranteed, it is reported in \cite{meng2019large} that $K$ is approximately proportional to $p$ in practice, in which case the computational cost for PPMM becomes $O(np^3+n\log(n)p)$.
Compared with the computational cost for the Sinkhorn algorithm, i.e., $O(n^2\log(n)p)$, PPMM has a lower order of the computational cost when $p\ll n$.
We illustrate Algorithm~\ref{alg:ALG2} in Fig.~\ref{fig:alg2}.
Although not covered in this section, the PPMM method can be easily extended to calculate the OTP, with minor modifications \cite{meng2019large}.

\section{Applications in Biomedical Research}

In this section, we present some cutting-edge applications of optimal transport methods in biomedical research.
We first present how optimal transport methods can be utilized to identify developmental trajectories of single cells \cite{schiebinger2019optimal}.
We then review a novel method for augmenting the single-cell RNA-seq data \cite{marouf2020realistic}.
The method utilizes the technique of generative adversarial networks (GAN), which is closely related to optimal transport methods, as we will discuss later.

\subsection{Identify Development Trajectories in Reprogramming}
The rapid development of single-cell RNA sequencing (scRNA-seq) technologies has enabled researchers to identify cell types in a population.
These technologies help researchers to answer some fundamental questions in biology, including how individual cells differentiate to form tissues, how tissues function in a coordinated and flexible fashion, and which gene regulatory mechanisms support these processes \cite{tanay2017scaling}.

Although sc-RNA-seq technologies have been opening up new ways to tackle the aforementioned questions, other questions remain.
Since these technologies require to destroy cells in the course of sequencing their gene expression profiles, researchers cannot follow the expression of the same cell across time.
Without further analysis, researchers thus are not able to answer the questions like what was the origin of certain cells at earlier stages and their possible fates at later stages; what and how regulatory programs control the dynamics of cells?
To answer these questions, one natural solution is to develop computational tools to connect the cells within different time points into a continuous cell trajectory.
In other words, although different cells are recorded in each time point, for each cell, the goal is to identify the ones that are analogous to its origins and its fates in earlier stages and late stages, respectively. 
A large number of methods have been developed to achieve this goal; see \cite{kester2018single,saelens2019comparison,farrell2018single,fischer2019inferring} and the reference therein.

A novel approach was proposed in \cite{tanay2017scaling} to reconstruct cell trajectories.
They model the differentiating population of cells as a stochastic process on a high-dimensional expression space. 
Recall that different cells are recorded independently at different time points. 
Consequently, the unknown fact to the researchers is the joint distribution of expression of the unobserved cells between different pairs of time points.
To infer how the differentiation process evolves over time, the authors assume the expression of each cell changes within a relatively small range over short periods.
Based on such an assumption, one thus can infer the differentiation process though optimal transport methods, which naturally gives the transport map between two distributions, respecting to two time points, with the minimum transport cost.

\begin{figure}[h]
    \begin{center}
        \begin{tabular}{cc}
            \includegraphics[width=0.95\textwidth]{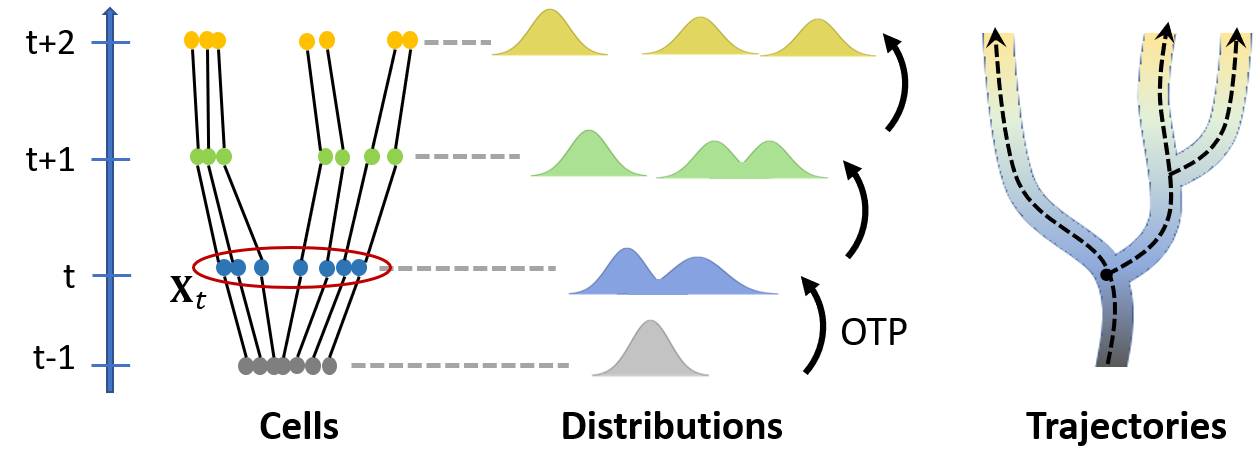}
        \end{tabular}
        \caption{Illustration for cell trajectories along time. Left: cells at each time point. Middle: OPT between distributions over cells at each time point. Right: cell trajectories based on OPT. }\label{fig:trajectory}
    \end{center}
    \vspace*{-0.2in}
\end{figure}

Figure~\ref{fig:trajectory} illustrates an idea to search for the ``cell trajectories''.
For gene expression $\X_t$ of any set of cells at time $t$, it can be transported to a later time point $t+1$ according to OTP from the distribution over $\X_t$ to the distribution over the cells at time $t+1$.
Analogously, $\X_t$ can be transported from a former time point $t-1$ by back-winding the OPT from the distribution over $\X_t$ to the distribution over the cells at time $t-1$ (The left and middle panels in Fig.~\ref{fig:trajectory}). 
The trajectory combines the transportation between any two neiboring time points (The right panel in Fig.~\ref{fig:trajectory}). Thus, OTP helps to infer the differentiation process of cells the at any time along the trajectory. 

The authors in \cite{tanay2017scaling} used optimal transport methods to calculate the differentiation process between consecutive time points and then compose all the transport maps together to obtain the cell trajectories over long time-intervals.
The authors also considered unbalanced transport \cite{chizat2018scaling} for modeling cellular proliferation, i.e., cell growth and death.
Analyzing around 315,000 cell profiles sampled densely across 18 days, the authors found reprogramming unleashes a much wider range of developmental programs and subprograms than previously characterized.

\subsection{Data Augmentation for Biomedical Data}

Recent advances in scRNA-seq technologies have enabled researchers to measure the expression of thousands of genes at the same time and to scrutinize the complex interactions in biological systems.
Despite wide applications, such technologies may fail to quantify all the complexity in biological systems in the cases when the number of observations is relatively small, due to economic or ethical considerations or simply because the sample size of available patients is low \cite{munafo2017manifesto}. 
The problem of a small sample size results in biased results since a small sample may not be a decent representative of the population.

Not only arising from biomedical research, such a problem  also arises from the research in various fields, including computer vision and deep learning, which require considerable quantity and diversity of data during the training process \cite{lecun2015deep,goodfellow2016deep}. 
In these fields, data augmentation is a widely-applied strategy to alleviate the problem of small sample sizes, without actually collecting new data.
In computer vision, some elementary algorithms for data augmentation include cropping, rotating, and flipping; see \cite{shorten2019survey} for a survey.
These algorithms, however, may not be suitable for augmenting data in biomedical research.

Compared with these elementary algorithms, a more sophisticated approach for data augmentation is to use generative models, including generative adversarial nets (GAN) \cite{goodfellow2014generative}, the ``decoder'' network in variational autoencoders \cite{kingma2013auto}, among others.
Generative models aim to generate ``fake'' samples that are indistinguishable from the genuine ones.
The fake samples then can be used, alongside the genuine ones, in down-stream analysis to artificially increase sample sizes.
Generative models have been widely used for generating realistic images \cite{dosovitskiy2016generating,liu2017auto}, songs \cite{blaauw2016modeling,engel2017neural}, and videos \cite{liang2017dual,vondrick2016generating}.
Many variants of the GAN method have been proposed recently, and of particular interest is the Wasserstein GAN \cite{arjovsky2017wasserstein}, which utilizes the Wasserstein distance instead of the Jensen–Shannon divergence in the standard GAN for measuring the discrepancy between two samples.
The authors showed that the Wasserstein GAN yields a more stable training process compared with the standard GAN, since Wasserstein distance appears to be a more powerful metric than the Jensen–Shannon divergence in GAN.

Nowadays, GAN has been widely used for data augmentation in various biomedical research \cite{frid2018synthetic, frid2018gan,madani2018chest}.
Recently, \cite{marouf2020realistic} proposed a novel data augmentation method for scRNA-seq data. 
The proposed method, called single-cell GAN, is developed based on Wasserstein GAN.
The authors showed the proposed method  improves downstream analyses such as the detection of marker genes, the robustness and reliability of classifiers, and the assessment of novel analysis algorithms, resulting in the potential reduction of the number of animal experiments and costs.

Note that generative models are closely related to optimal transport methods.
Intuitively, a generative model is equivalent to finding a transport map from random noises with a simple distribution, e.g., Gaussian distribution or uniform distribution, to the underlying population distribution of the genuine sample.
Recent studies suggest optimal transport methods outperform the Wasserstein GAN for approximating probability measures in some special cases \cite{lei2019geometric,lei2020geometric}.
Consequently, researchers may consider using optimal transport methods instead of GAN models for data augmentation in biomedical research for potentially better performance.

\section*{Acknowledgment}
The authors would like to acknowledge the support from the U.S. National Science  Foundation under grants DMS-1903226,  DMS-1925066, the U.S. National Institute of Health under grant R01GM122080.


\bibliography{ref}
\bibliographystyle{abbrv}
\end{document}